# A Precisely Xtreme-Multi Channel Hybrid Approach For Roman Urdu Sentiment Analysis

**Faiza Memood[1], Muhammad Usman Ghani[2], Muhammad Ali Ibrahim[3], Rehab Shehzadi[4], Muhammad Nabeel Asim[5]**

# 1 Abstract

In order to accelerate the performance of various Natural Language Processing tasks for Roman Urdu, this paper for the very first time provides 3 neural word embeddings prepared using most widely used approaches namely Word2vec, FastText, and Glove. The integrity of generated neural word embeddings is evaluated using intrinsic and extrinsic evaluation approaches. Considering the lack of publicly available benchmark datasets, it provides a first-ever Roman Urdu dataset which consists of 3241 sentiments annotated against positive, negative and neutral classes. To provide benchmark baseline performance over the presented dataset, we adapt diverse machine learning (Support Vector Machine Logistic Regression, Naive Bayes), deep learning (convolutional neural network, recurrent neural network), and hybrid approaches. Effectiveness of generated neural word embeddings is evaluated by comparing the performance of machine and deep learning based methodologies using 7, and 5 distinct feature representation approaches respectively. Finally, it proposes a novel precisely extreme multi-channel hybrid methodology which outperforms state-of-the-art adapted machine and deep learning approaches by the figure of 9%, and 4% in terms of F1-score. Roman Urdu Sentiment Analysis, Pretrain word embeddings for Roman Urdu, Word2Vec, Glove, Fast-Text

# 2 Introduction

The trend of using social media platforms ( e.g Facebook, Twitter, Tumblr, Reddit) to communicate with family and friends, sharing the experiences, and opinions regarding a particular product, service, person, or organization has become exceptionally common. According to a recent report published by marketers at the official MediaKix platform [1], people spend way more time over social media sites than they usually do on drinking, eating, and combined socializing. Likewise, according to SmartInsights [2] survey, people manage to publish 3.3 million posts on Facebook, 4.5 million tweets on Twitter within a minute. These compelling statistics of being addicted to social media platforms are elevating further with the speed of light. Considering the extensive usage of social media sites, extracting and analyzing user reviews related to a certain event, issue, product, service, organization or celebrity, a dedicated task known as Sentiment Analysis has become a promising area of Natural Language Processing (NLP). One of the most beckoning reasons for extensively leveraging sentiment analysis is that it largely assists the companies to comprehend consumer needs and formulate imperative modifications in marketing and business strategies to enhance user experience [1][2]. Recent advancements in machine and deep learning based sentiment analysis methodologies have significantly uplifted the performance of multifarious business intelligence [3][4][5], scientific [6][7][8][9], and academic applications [10][11][12] by acquiring noteworthy insights, and substantially raising the product or service standards.

---
[1]https://mediakix.com/blog/how-much-time-is-spent-on-social-media-lifetime/gs.x0iGr30
[2]https://www.smartinsights.com/internet-marketing-statistics/happens-online-60-seconds/

There exists a substantial number of symposiums, workshops, and conferences which primarily focus on the discovery and smart processing of sentiments extracted from diverse social media platforms. A few such renowned resources are Sentiment Analysis Symposium (SAS)[3], Workshop on Computational Approaches to Subjectivity, Sentiment and Social Media Analysis (WASSA)[4], Opinion Mining, Summarization and Diversification (WISDOM)[5], ACM conference for Knowledge Discovery and Data Mining (SIGKOD)[6]. Such platforms provide an international forum for worldwide researchers to share the latest findings related to social data mining and their potential applications in both academia and industrial regions. These tracks also facilitate benchmark corpora for various languages including English, Chinese, German and Arabic to accelerate sentiment analysis research. The availability of such rich resources has largely aided the researchers to perform a comparative analysis of diverse machine and deep learning methodologies and to assess the effectiveness of enhanced novel methodologies. Evidently, this progress has led the emergence of jaw-dropping applications for these rich resourced languages which are capable to perform sentiment classification in real-time such as Nexmo [7], intent detection like LiveIntent[8], emotion identification [13], emotion classification [14], constructing user interests profile [15][16], and user reaction categorization [17].

In contrast, South-Asian languages specifically Roman Urdu which has more than 100 million speakers worldwide is considered an under-resourced language in this regard. Few conferences like International Joint Conference of Natural Language Processing (IJCNLP) [9] has provided linguistic resources for Asian languages to support the processing of diverse tasks involving named entity recognition (NER), language parsing, phonology, morphology, and word segmentation [10]. However, existing con-ferences or special issue tracks do not provide resources for Roman Urdu. This has not only substantially impeded the development of novel sentiment classification methodologies, but also hampered in-depth performance analysis that could have been performed through adapting state-of-the-art sentiment classification methodologies. Resultantly, no application exists for Roman Urdu which can perform sentiment analysis in real time. In order to support the development of NLP applications for Roman Urdu, considering the deficiency of publicly available Roman Urdu dataset, the paper in hand presents the first-ever publicly available benchmark Roman Urdu sentiment dataset.

---
[3] http://2018.sentimentsymposium.com/
[4] https://wt-public.emm4u.eu/wassa2019/index.htm
[5] https://www.aclweb.org/portal/content/cfp-7th-kdd-workshop-issues-sentiment-discovery-and-opinion-mining-wisdom18
[6] https://www.kdd.org/proceedings/view/kdd-17-proceedings-of-the-23rd-acm-sigkdd-international-conference-on-knowl
[7] https://www.nexmo.com/use-cases/real-time-sentiment-analysis
[8] https://www.liveperson.com/products/liveintent/?utm$_s$ource = Field%20Service%20News
[9] https://www.emnlp-ijcnlp2019.org/
[10] http://www.afnlp.org/wp/?page$_i$d = 106

A very limited sentiment analysis work exists for Roman Urdu which can be classified into lexicon based, machine learning, and deep learning based approaches. Lexicon based approaches have low applicability over unseen data, and machine learning based approaches predominantly use bag-of-words based feature representation approaches which face the problem of data sparsity. Whereas, deep learning approaches utilize less effective feature representation approaches like one-hot encoding or randomly initialized neural word embeddings as there does not exist any pre-trained neural word embeddings for Roman Urdu.

Considering the promising performance produced by neural word embeddings (Word2vec[18], FastText [19], and Glove [20]) in variety of NLP tasks including hierarchical text categorization [21], multi-class text document classification [22][23], investigation of gender roles [24], non-relevant post detection [25], topic modelling [26], automated sarcasm detection [27], synonym extraction [28], automated enrichment of lexicons for misogyny detection [29], sentiment analysis [30], automated text summarization [31], text clustering [32], measuring emotional polarity from debates [33], recommendation system [34], the paper in hand for the very first time provides Word2vec[18], FastText [19], and Glove [20] embeddings for Roman Urdu. These pre-trained embeddings can be used to enhance the performance of diverse deep learning based Roman Urdu processing tasks. To assess the integrity of generated embeddings, evaluation is performed in two different manners. Firstly, the degree of word relatedness is performed using t-distributed stochastic neighbor embedding (t-SNE) [35]. Secondly, in order to assess the degree of separation among distinct sentiments, document embeddings are prepared in a supervised manner and segregation among the clusters of document embeddings is visualized using t-SNE [35]. In addition, the more traditional approach is used in which embeddings are evaluated through a downstream task known as Sentiment Analysis.

To provide benchmark performance for the task of sentiment analysis on the newly developed dataset, we have performed extensive experimentation by adapting 3 machine learning based methodologies (Support Vector Machine (SVM) [36], Logistic Regression (LR) [37], Naive Bayes (NB) [38]), and 8 deep learning based methodologies (convolutional neural network (CNN) [39] [40], recurrent neural network (RNN) [41], and Hybrid approach [42]). For adapted machine learning based methodologies, we compare the performance of 7 different feature representation approaches (TF-IDF [43], Word2vec [18], FastText [19], Glove [20], Doc2vec, Doc_FastText, Doc_Glove). Whereas, for adapted deep learning based methodologies, we compare the performance of 5 different feature representation approaches (TF-IDF [43], randomly initialized word embeddings, Word2vec [18], FastText [19], Glove [20]). Finally, we present a novel precisely extreme-multi-channel hybrid methodology for Roman Urdu sentiment analysis. The proposed methodology outshines adapted machine learning based methodologies by the figure of 7%, 10%, 6%, 9%, and deep learning methodologies by the figure of 3%, 4%, 5%, 4% in terms of accuracy, precision, recall, and F1-score. The contribution of this paper can be summarized as:

1. It provides pre-trained neural word embeddings of three most widely used approaches Word2vec[18], FastText [19], and Glove [20] prepared over a gigantic corpus containing 6.2 million Roman Urdu text.

2. It extensively evaluates the integrity of neural word embeddings using intrinsic and extrinsic evaluation measures.

3. It provides a publicly available sentiment analysis dataset containing 9006 features, and 3241 Roman Urdu sentiments to eliminate a major hindrance in the evaluation of sentiment analysis approaches.

4. To provide benchmark performance, we perform extensive experimentation on newly developed dataset with 4 different evaluation measures by adapting 3 machine learning based methodologies, and 8 deep learning based methodologies. Sentiment analysis as a downstream task is performed using adapted machine and deep learning based methodologies with 7, and 5 unique feature representation approaches respectively.

5. Finally, we propose a novel precisely extreme multi-channel hybrid methodology which significantly outperforms state-of-the-art machine and deep learning based classification methodologies across 4 different evaluation metrics.

The rest of the paper first critically analyzes the previous work solely related to Roman Urdu sentiment analysis. Then, it deep dives into the generation of corpora, and neural word embeddings followed by proposed and adapted methodologies along with evaluation metrics. Afterward, it briefly discusses experimental setup before comparing the results of adapted machine and deep learning methodologies with the proposed methodology. Finally, it highlights the key findings of experimentation and gives future directions.

## 3 Roman Urdu Sentiment Analysis

Sentiment analysis is the core building block behind the development of more appealing marketing and branding strategies including accelerating business sales through dynamic pricing and enhancing user experience through efficient technical support [1][2]. Compared to other rich-resourced languages, a limited amount of work has been performed for Roman Urdu sentiment analysis, which is summarized below.

In 2019, Ayesha et al [44] crawled several websites to prepare a Roman Urdu dataset containing opinions about various products and services. They employed three machine learning classifiers including Naive Bayes, Support Vector Machine, and Logistic Regression with Stochastic Gradient Descent to assess the polarity of extracted opinions. Through experimentation, they found that SVM managed to outperform other classifiers. Bilal et al [45] first extracted 300 positive and negative opinions expressed in Roman Urdu, and English from a blog. Afterwards, they performed sentiment analysis using three diverse machine learning classifiers including Naive Bayes, KNN, and Decision Tree. Experimental results showed that Naive Bayes overshadowed the performance of KNN, and Decision Tree in terms of four evaluation metrics accuracy, precision, recall, and F1-score.

Khan et al [46] prepared a dataset of reviews by scrapping several automobile websites and classifying them against positive and negative classes. Experimentation for Roman Urdu text classification was performed using Multinomial Naive Bayes, Random Forest, Decision Tree, SVM, kNN, Bagging, and very simple multi-layer perceptron network. Authors found that Multinomial Naive Bayes managed to attain the highest accuracy, precision, recall, and F1-score amongst all classifiers. Mehmood et al. [47] presented a sentiment analysis end to end system for Roman Urdu. They prepared a dataset of 779 reviews belonging to five domains including Mobile phones, Movies, Miscellaneous, Politics, and Drama. They considered n-gram features and experimented with five machine learning classifiers namely Logistic Regression (LR), and Naive Bayes (NB), kNN, SVM, and Decision Tree. Amongst all, two classifiers Logistic Regression (LR), and Naive Bayes (NB) marked competitive performance.

Arif et al. [48] carried the task of sentiment analysis over Roman Urdu corpus which was prepared by translating existing Hotel reviews expressed in the English language. For experimentation, authors utilized 3 feature representation approaches (TF, TF-IDF, Hashingvectorizer), and 3 feature selection approaches (Chi-Squared, IG, MI) and 10 classifiers including SVM, kNN, Decision Tree, Passive Aggressive, Ensemble classifier, Perceptron, SGD, Naive Bayes, Ridge classifier, and nearest centroid. Amongst all machine learning based classifiers, SVM produced more promising performance with all feature representation and selection approaches.

Hasan et al. [49] adopted a hybrid methodology in which they experimented with diverse lexicons and machine learning classifiers for election sentiments analysis. Authors performed experimentation with three lexicons including SentiWordNet [11], TextBlob [50], and Wordnet with Word Sense Disambiguation (W-WSD) [12], and two machine learning classifiers (SVM, NB). They reported that WordNet and TextBlob were highly accurate in word sense disambiguation and largely assisted the classifier to detect polarity in political reviews. Mehmood et al. [51] presented a novel feature representation approach namely "Discriminative Feature Spamming" for Roman Urdu sentiment analysis. They compared the performance of the presented approach with TF, Binary Weighting, TF-IDF with word and character level features using Naive Bayes, Logistic Regression, majority voting, weighted voting, and multi-layer perceptron. They reported that the proposed feature representation approach significantly raised the performance of all classifiers. Amongst all, weighted voting algorithm marked the best performance.

Noor et al. [52] collected reviews from an e-commerce Pakistan site namely Daraz [13] and classified into positive, negative, and neutral classes. Authors utilized bag-of-words based model for feature extraction which were later fed into Support Vector

---

[11] http://sentiwordnet.isti.cnr.it/

[12] https://github.com/kevincobain2000/sentiment$_classifier$

[13] https://www.daraz.pk/

Machine (SVM) for the task of sentiment categorization. Mehmood et al. [53] established a corpus of belonging to six diverse domains. For experimentation, authors used word level features, character level features, and union of both. They found that they managed to reduce the error by the figure of 12% from baseline (80%).

On the other hand, considering the promising performance produced by Recurrent Neural Networks (RNN) in multifarious Natural Language Processing tasks, Ghulam et al. [54] utilized Long-short time memory model (LSTM [55]). Through experimentation, authors found that LSTM [55] significantly outperformed machine learning based approaches.

In a nutshell, preliminary work makes use of either bag-of-words based approaches, or randomly initialized neural word embeddings for feature representation. While bag-of-words based approaches face the problem of data sparsity, randomly initialized word embeddings do not capture the semantics of language and fail to overshadow the performance of pre-trained neural word embeddings. Moreover, mostly state-of-the-art work employs machine learning based methodologies, only one researcher has utilized deep learning based approach for the task of Roman Urdu sentiment analysis. Considering the extensive usage of SVM, LR, and NB in sentiment analysis literature and their effectiveness for text classification, to evaluate the integrity of proposed methodology, we have performed experimentation over presented dataset with only these three classifiers. As our main focus is to evaluate diverse deep learning methodologies for the task of Roman Urdu sentiment analysis.

## 4 Materials And Methods

This section briefly describes the characteristics of Roman Urdu corpus and three deep learning approaches used for the generation of neural word embeddings. It discusses the specificities of developed benchmark dataset for the task of Roman Urdu sentiment analysis. Finally, it deep dives into proposed novel methodology followed by adapted methodologies, and evaluation metrics used for the performance comparison.

### 4.1 Roman Urdu Corpus For Embedding Generation

In order to generate neural word embeddings for the effective representation of Roman Urdu sentiments, we have prepared an enormous corpus containing 6.2 million Roman Urdu text. Entire corpus is crawled from social media handle "Twitter" [14], and a mobile review domestic website namely "WhatMobile" [15]. Extensive use of social media and brand review websites are producing a totally new style of written text usually referred as MicroText. MicroText itself is extremely noisy, however for Roman Urdu, it is even more complicated as it may contain special symbols, relaxed spellings (e.g acha, achaa, achha, aacha for the word good), out of vocabulary (OOV) words like emotional stress (aaaaala, for the word too good) phonetic spellings such as (yr is the slang of yaar (Friend)). Deep comprehension of microtext of a certain language is mandatory for effectively processing it.

---
[14] https://twitter.com/
[15] https://www.whatmobile.com.pk/

In order to effectively capture and represent user sentiments, we have normalized the microtext of Roman Urdu by modifying the linguistic rules given by Zareen et al. [56] and defining 100 new rules. Mainly, all defined linguistic rules are based on word phonetics. To illustrate this point, all words including "Kesi" "kesy", "kesyy", "kesiy", "kesii" are transformed into "Kese" considering the phonetics of word ending characters (e.g i, y). Nevertheless, as Roman Urdu is a linguistically rich and morphologically complex language, thus defined rules manage to normalize only a few words.

### 4.2 Neural Word Embedding Space Construction

Pre-trained neural word embeddings have brought Natural Language Processing a long way by largely assisting deep learning methodologies to attain promising results over diversified NLP tasks [57][58]. The impact brought by continuous distributed word vectors [57] is greatly similar to the impact produced by pre-trained ImageNet models for multifarious computer vision tasks [59] [60]. There exists a variety of domain-specific and cross-domain pre-trained neural word embeddings for several rich resourced languages involving English, Chinese, German, and Arabic [16], however, there does not exist any kind of pre-trained neural word embeddings for Roman Urdu.

Neural word embeddings are even more essential for convoluted languages like Roman Urdu where a great number of variations are possible for every word [56]. For instance, the word beautiful can be expressed by so many ways in Roman Urdu such as "khubsoorat", 'khbsrat''', 'khoobsurat''','khobsurt''', and many more. Generally, these embeddings are compendious word meaning vectors obtained by training deep neural networks in an unsupervised manner to solve a certain task. More specifically the task is to predict a missing word by processing a word sequence containing the surrounding words. Neural network hidden layer determines the meaning of every word on the basis of context it has gone through and generates condensed optimal representation [61]. These embeddings are not only dense, much smaller, and memory-efficient but also effectively capture word associations including word synonyms, and antonyms [62] such as Aadmi-Shakhs, Larka-Larki, etc. Diverse deep learning methodologies used for the generation of neural word embeddings are discussed in subsequent sections.

#### 4.2.1 Word2Vec

Word2vec [18] is considered a predictive neural word embedding model that learns the representations by predicting the target word from the surrounding words. Mainly, Word2vec [18] has two architectures that can be used to learn distributed representations of corpus words namely continuous bag-of-words (CBOW), and continuous skip-gram (CSG). Continuous bag-of-words prediction does not affect the order of surrounding words as the model makes use of the current word to infer the window of context words. On the other hand, continuous skip-gram assigns more weight to nearby surrounding words as compared to far away context words and the model predicts the central word using a weighted window of surrounding words. Word2vec [18] both architectures only use local context and learns unified vector representation for each word, however, there is a strong possibility that a word may appear in multiple dissimilar contexts.

---
[16]https://fasttext.cc/docs/en/crawl-vectors.html

### 4.2.2 Glove

As Word2vec [18] does not take global context into account, thus Glove [20] neural word embeddings came into picture. Glove embeddings make use of the same intuition behind distributional embeddings of a co-occurring matrix, the only difference is that it utilizes a neural network to decompose a co-occurring matrix into compact word vectors. Glove [20] word vectors have shown better performance than Word2vec [18] in word analogy tasks as Glove [20] adds more meaning into neural word embeddings by taking the relationship among word pair to word pain into account. In addition, Glove [20] assigns lower weights to highly frequent word pairs including "a", "the", etc. However, as the model is based on a co-occurrence matrix, hence Glove [20] requires a huge amount of memory for storage. Also, changing hyperparameters closely related to the co-occurrence matrix, one needs to reconstruct the entire matrix again which will consume a hefty amount of time.

### 4.2.3 FastText

In order to effectively learn the representation of out of vocabulary (OOV) words, a common problem faced by both Word2vec [18], and Glove [20], FastText [19] just like Word2vec [18] learns the vector representation of each word and also the n-grams located within every word. Afterwards, representation values are averaged to create a unified vector at every training step. Although these embeddings are computationally more expensive than Word2vec [18], and Glove [20], however it permits the neural word embeddings to encode notable sub-word information. FastText neural word embeddings are far more accurate than Word2vec [18] when evaluated using several measures.

## 4.3 Benchmark Dataset: DSL Roman-Urdu Sentiments

For the evaluation of neural word embeddings in terms of their ability to capture overall concept of a document, and to perform Roman Urdu sentiment analysis, considering the unavailability of dataset, we present a publicly available benchmark dataset namely "DSL Roman-Urdu Sentiments". DSL Roman-Urdu Sentiments corpus consists of 3241 mobile related sentiments manually annotated against positive, negative and neutral intents. Entire dataset is crawled from mobile review website namely WhatMobile [17]. Pre-processing of the corpus is performed in a same manner as applied for other corpus used for the generation of neural word embeddings (discussed in section 4.1)

## 4.4 Proposed Methodology

It was initially considered that convolutional neural networks (CNN) generally perform better only for computer vision tasks by recognizing notable patterns across the space

---
[17] https://www.whatmobile.com.pk/

[59] [60]. Whereas their counterparts recurrent neural networks (RNN) extract patterns across timestamps through the chain of neural network blocks and are more appropriate to handle textual data [63]. In last decade, researchers have shown the effectiveness of RNNs for the task of sentiment classification [64] machine translation [65], handwriting recognition [66], language modelling [67][68] [69] and question answering [70]. However, in recent times, researchers have proved that CNNs are capable to outshine RNN and its variants (LSTM [55], GRU [71]) over variety of NLP tasks such as language modelling [72], long sentences categorization [73], relation classification[74], and answer selection [75].

Because of these jaw-dropping findings and lack of in-depth comparative analysis of CNN, and RNN for a variety of NLP tasks, selecting appropriate model (CNN or RNN) for hand on NLP task has become a point of contention. Building on this, trend of taking the advantage of both architectures (CNN, RNN) in the form a hybrid methodology has significantly elevated. Inspiring from the performance improvement produced by hybrid methodologies [74][75] especially for sentiment analysis tasks [76][77][78], in order to reap the benefits of both CNN, and RNN along with a variety of pre-trained neural word embeddings, we have proposed a precisely extreme multi-channel hybrid methodology. The proposed methodology makes use of uni and bi-directional GRUs [71], and CNN layers.

Chung et al [79], and Jozefowicz et al. [80] empirical evaluations have reported that both GRU [71] and LSTM [55] produce a comparable performance in several NLP tasks and one can not be considered better than other because tuning few hyperparameters such as layer size are primarily deriving the performance. Considering GRU [71] has fewer parameters, are more memory and time-efficient as compared to other architectures specifically used to handle sequential data (e.g LSTM [55]), and requires less training data to generalize well on unseen data, we have utilized GRU [71] in proposed methodology for hand on sentiment analysis task [81], Before dwelling into the architecture of proposed methodology, lets first have a look at basic building blocks of GRU [71].

Turning towards how GRUs [71] are utilized in proposed precisely extreme multi-channel hybrid methodology, Figure 1 shows the graphical representation of proposed methodology.

As is illustrated by the Figure 1, in order to reap the benefits of 3 neural word embeddings, we have used a precisely multi-channel strategy where proposed model makes use of Word2vec [18] at first channel, FastText at second channel [19], and Glove [20] at third channel. We have utilized 3 uni-directional Gated Recurrent Units (GRUs [71]) in every channel. Central uni-directional GRU [71] attains the representation of current word, leftmost, and rightmost uni-directional GRUs [71] acquire the representation of left and right context word of current word using respective embedding layer. To avoid overfitting the model, embeddings are kept static for leftmost and central uni-directional GRU [71], whereas embeddings of rightmost uni-directional GRU [71] are further fine tuned. Afterwards, in order to create a unified representation based on semantic similarity, representation of three channels utilizing Word2vec [18], FastText [19], and Glove [20] are concatenated for every corpus word. Considering the effectiveness of RNN for learning long range dependencies, and CNN for the acquisition of promising features, yielded unified representation is first passed to a Bi-directional GRU [71] which better extracts contextual information. Afterwards, most discriminative features are extracted through max pooling and passed to a fully connected layer.

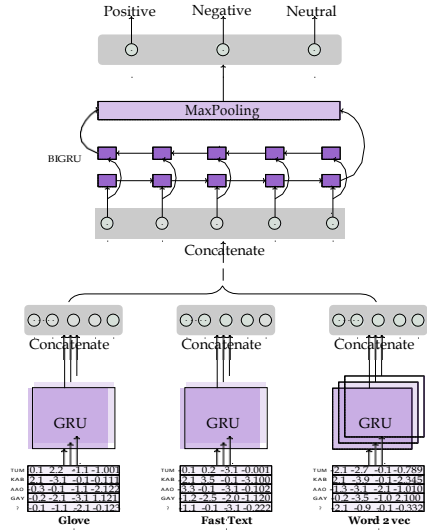

Figure 1: Proposed Precisely Extreme Multi-Channel Hybrid Methodology

### 4.4.1 Baseline

Although few researchers have carried the task of Roman Urdu sentiment analysis, however not even a single dataset is publicly available. Here we have developed a Roman Urdu sentiment dataset to carry extrinsic evaluation of pre-trained neural word embeddings through a downstream sentiment analysis task. In order to compare the performance of proposed precisely multi-channel hybrid methodology, we have adapted diverse machine and deep learning methodologies, detail of which are briefly discussed below.

Considering the promising performance of Support Vector Machine (SVM) [36], Logistic Regression (LR) [37], and Naive Bayes (NB) [38] as described by the literature, to get the benchmark performance over developed dataset, we have performed extensive experimentation with these classifier using 7 different feature representation approaches. Different feature representation approaches are utilized to compare the performance of pre-trained neural word and document embeddings with trivial TF-IDF [43] feature representation approach.

SVM is considered a linear but non-probabilistic classifier which maps every instance in a multi-dimensional Cartesian plane and determines the most distant hyperplane which best segregates class boundaries. It comes under the hood of *Discrimi-* *native Classifiers* and largely utilized for categorization [82], and anomaly detection tasks [83]. Whereas NB makes use of probability theory and bayes theorem for class inference. NB classifier comes under the umbrella of *Generative Classifiers* and mostly use for spam detection [84], and text document classification [85]. Likewise, LR is another probabilistic classifier, an approach borrowed from the domain of statistics. LR utilizes maximum likelihood estimation algorithm to alleviate the error in predicted probabilities. LR is considered a good baseline which is extensively used to estimate the performance of complex algorithms.

On the other hand, as deep learning methodologies are broadly classified into CNN, RNN, and Hybrid approaches, thus we have adapted 2 CNN based, 3 RNN based, and 3 hybrid methodologies to get benchmark performance over all three kinds of deep learning approaches for Roman Urdu sentiment analysis.

For CNN based methodologies, we adapt a CNN model presented by Kalchbrenner et al. [39] for sentiment analysis task. Authors for the very first time utilized wide convolutions. They reported that in case of large filter size, words residing at edges of certain document are usually neglected during convolution. Considering the fact that a discriminative feature may present anywhere in the document, by the use of wide convolutions, authors made sure that every word is equally participating in convolution.

As our proposed methodology is based on multiple channels, hence to perform a fair comparison, we have also adapted a multi-channel CNN model proposed by Yoon Kim [40] for the task of sentiment classification. They for the very first time presented a multi-channel approach for textual data which utilized different feature representation approaches at different channels by making few channels static throughout to avoid overfitting. They reported that CNN model shows better performance when the embedding layer is fed with pre-trained neural word embeddings which are further fine-tuned during training. Their model outshined 14 diverse classification methodologies [40].

To prove the effectiveness of proposed precisely extreme multi-channel hybrid methodology in the extraction of local and global features for sentiment analysis, we adapted an LSTM [55] model presented by Xuangjing et al. [41] with same intuition. Their model utilized a cache mechanism to segregate internal memory into multiple unique groups having diverse memory cycles by squeezing the forgetting rates. Resultantly, it did only help the model to acquire global, and local sentiment information but also largely assisted the model to converge faster as the gradient got stable during back propagation.

Considering our proposed sentiment analysis methodology is hybrid in nature, we adapted a hybrid model based on CNN, and LSTM [55] presented by Chen et al. [42] for the task of text categorization. They utilized pre-trained neural word embeddings for feature representation, CNN for feature extraction followed by LSTM [55] layer. Authors reported that pre-trained semantic similarity based word vectors contained local features of every word and largely assisted CNN to acquire global features of every word. Both features were effectively utilized by LSTM [55] to estimate the combination of labels for a given instance.

While adapting discussed CNN, RNN, and Hybrid classification methodologies for Roman Urdu sentiment analysis, we have not only experimented with TF-IDF [43], randomly initialized embeddings, Word2vec [18], FastText [19], and Glove [20] word vectors but also experimented with all three sequence processing architectures includ-ing RNN, LSTM [55], and GRU [71].

## 4.5 Evaluation Measures

This section briefly discusses the evaluation measures used to compare the performance of proposed precisely extreme multi-channel hybrid methodology with adapted machine and deep learning based methodologies. All utilized multi-class evaluation metrics are described below:

### 4.5.1 Accuracy

Accuracy [86] is the proportion of correctly predicted samples to all types of predictions made by the model. Mathematically, it is defined as:

$$Accuracy(A) = \frac{tp + tn}{tp + fp + tn + fn}$$

### 4.5.2 Precision

Precision [86] measures how many samples that are predicted as positive by the model, actually belong to positive class. It can be defined in the following way:

$$Precision(P) = \frac{tp}{tp + fp}$$

### 4.5.3 Recall

Recall [86] estimates what proportion of samples that actually belong to the positive class, are correctly predicted as positive by the model. Mathematically, it is written in the following way:

$$Recall(R) = \frac{tp}{tp + tn}$$

### 4.5.4 F1 Score

F1 Score [86] is computed by taking the harmonic mean of precision, and recall. It is defined in the following way:

$$F1\_Score(F) = \frac{2 * P * R}{P + R}$$

## 5 Experimental Setup And Results

Roman Urdu sentiments for both annotated and non-annotated experimental datasets are crawled and parsed using BeautifulSoup [18]. Neural word embeddings are learned from an enormous non-annotated corpus using Gensim [19]. Word2vec continuous bag-of-words [18], FastText [19], and Glove [20] embeddings are created with 200 dimensions by training the model for 20 epochs. While for Word2vec [18], and FastText [19] maximum distance among the analyzed set of words within same sentence is 10 as compared Glove where the window size is 15. For all three neural word embedding approaches, words having a frequency lower than 5 are ignored. To explore the performance impact of generated embeddings, for a downstream sentiment analysis task, while machine learning based adapted methodologies are implemented using Scikit-Learn [20], deep learning based methodologies are implemented using Keras API [21].

---
[18]https://pypi.org/project/beautifulsoup4/
[19]https://pypi.org/project/gensim/

#### 5.0.1 Training Process

For Roman Urdu sentiment analysis, we have split the developed dataset into train, validation and test sets containing 60%, 10%, and 30% of corpus instances. Furthermore, we have used rMSprop [87] as an optimizer with learning rate of 0.01. Categorical cross entropy [88] is used to back propagate the loss. We have trained the model for 50 epoch with the patience of 5. Through early stopping, best performing model is saved and used during the evaluation of Roman Urdu sentiment analysis task.

### 5.1 Results

This section briefly describes the performance of proposed and adapted methodologies. The performance of machine and deep learning based adapted methodologies is assessed across 4 evaluation metrics by leveraging 7, and 5 different feature representation approaches.

| Algorithms | Embedding | Evaluation Measures | | | |
|---|---|---|---|---|---|
| | | Accuracy | Precision | Recall | F1_core |
| Our Model | 3(W2V+GloVe+FT) (static), 9 GRU | 0.7708 | 0.7634 | 0.7917 | 0.7581 |
| | 3(W2V+GloVe+FT) (static + non-static), 9 GRU [71] | 0.8099 | 0.7812 | 0.8136 | 0.7911 |
| | 3(W2V+GloVe+FT) (static), 9 GRU + Bi-GRU | 0.8243 | 0.8003 | 0.8006 | 0.80 |
| | 3(W2V+GloVe+FT) (static+non-static), 9 GRU + Bi-GRU | **0.8417** | **0.8168** | **0.8284** | **0.8221** |

Table 1: Performance Comparison of Proposed And Adapted Deep Learning Based Approaches Using Neural Word Embeddings And Bag-of-Words Based Feature Representation Approach

## 6 Conclusion

This paper achieves important landmarks in regard of Roman Urdu sentiment analysis. It provides a public benchmark sentiment analysis dataset along with 3 distinct pre-trained neural word embeddings. It also rigorously evaluates the performance impact of generated neural word embeddings, and bag-of-words based feature representation approaches by adapting a variety of machine and deep learning methodologies. While most machine learning methodologies perform better with TF-IDF, adapted deep learning methodologies mark superior performance with word2vec. Although hybrid approach and GRU outperform best performing SVM, and other adapted deep learning methodologies by a decent margin. However, proposed precisely extreme multi-channel hybrid methodology significantly outperforms machine and deep learning methodologies. It harvests the advantages of 3 pre-trained neural word embeddings through multiple channels for effective representation, bi-directional GRU for optimal contextual information, and CNN for the acquisition of extremely discriminative features. An impressive future line of current work would be investigating the performance impact of generated neural word embeddings for other NLP tasks such as machine translation, and cyber bullying detection.

---
[20]https://scikit-learn.org/stable/
[21]https://keras.io/